\documentclass[runningheads]{llncs}
\usepackage{graphicx}
\usepackage{amsmath,amssymb} % define this before the line numbering.
\usepackage{subfigure}
\usepackage{booktabs}
\usepackage{multirow}
\usepackage{tablefootnote}
\begin{document}
\title{Extreme Network Compression via Filter Group Approximation}
% Replace with your title

\titlerunning{Extreme Network Compression via Filter Group Approximation}
% Replace with a meaningful short version of your title
%
\author{Bo Peng \and
Wenming Tan \and
Zheyang Li \and
Shun Zhang \and
Di Xie \and
Shiliang Pu }
%
%Please write out author names in full in the paper, i.e. full given and family names.
%If any authors have names that can be parsed into FirstName LastName in multiple ways, please include the correct parsing, in a comment to the volume editors:
%\index{Lastnames, Firstnames}
%(Do not uncomment it, because you may introduce extra index items if you do that, we will use scripts for introducing index entries...)
\authorrunning{Bo Peng, Wenming Tan, Zheyang Li, Shun Zhang, Di Xie, Shiliang Pu}
% Replace with shorter version of the author list. If there are more authors than fits a line, please use A. Author et al.
%

\institute{Hikvision Research Institute, Hangzhou, China\\
\email{\{pengbo7,tanwenming,lizheyang,zhangshun7,xiedi,pushiliang\}@hikvision.com}}
\maketitle              % typeset the header of the contribution
\begin{abstract}
In this paper we propose a novel decomposition method based on filter group approximation, which can significantly reduce the redundancy of deep convolutional neural networks (CNNs) while maintaining the majority of feature representation. Unlike other low-rank decomposition algorithms which operate on spatial or channel dimension of filters, our proposed method mainly focuses on exploiting the filter group structure for each layer. For several commonly used CNN models, including VGG and ResNet, our method can reduce over 80\% floating-point operations (FLOPs) with less accuracy drop than state-of-the-art methods on various image classification datasets. Besides, experiments demonstrate that our method is conducive to alleviating degeneracy of the compressed network, which hurts the convergence and performance of the network.

\end{abstract}
\section{Introduction}
\label{Introduction}

In recent years, CNNs have achieved great success on several computer vision tasks, such as image classification \cite{Krizhevsky2012ImageNet}, object detection \cite{Ren2017Faster}, instance segmentation \cite{Noh2016Learning} and many others. However, deep neural networks with high performance also suffer from a huge amount of computation cost, restricting applications of these networks on the resource-constrained devices. One of the classic networks, VGG16 \cite{Simonyan2014Very} with 130 million parameters needs more than 30 billion FLOPs to classify a single 224$\times$224 image.
The heavy computation and memory cost can hardly be afforded by most of embedding-systems on real-time tasks. To address this problem, lots of studies have been proposed during last few years, including network compressing and accelerating \cite{Denton2014Exploiting,Han2015Deep,Li2016Pruning,Zhang2016Accelerating}, or directly designing more efficient architectures \cite{Howard2017MobileNets,Ioannou2016Deep,Xie2016Aggregated,Zhang2017ShuffleNet}.

\begin{figure}[tb]
\begin{center}
\centerline{\includegraphics[width=0.75\columnwidth]{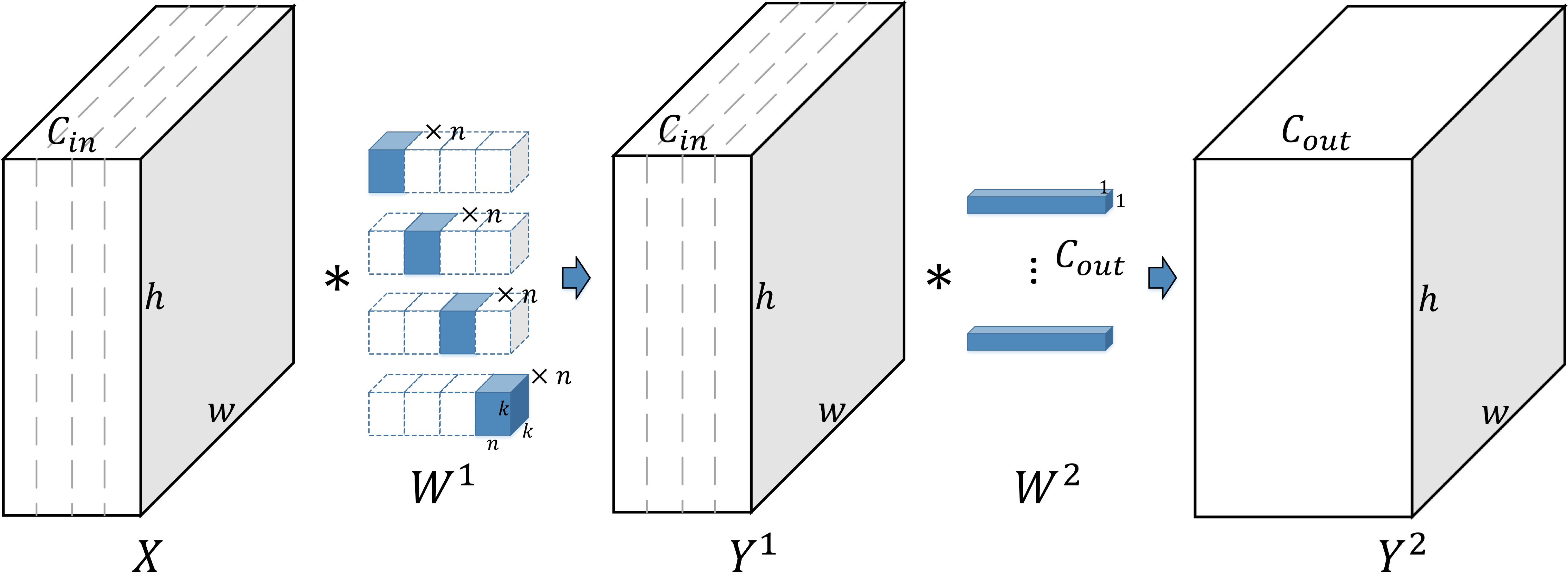}}
\caption{The filter group structure is a linear combination of $W^{1}$ and $W^{2}$. $X$ is the input feature with $C_{in}$ channels and $h \times w$ spatial size. $W^{1}$ is a filter group convolutional layer with each filter size of $n \times k \times k$, $W^{2}$ is a $1 \times 1$ convolutional layer. $Y^1$ and $Y^2$ are outputs of $W^{1}$ and $W^{2}$ respectively }
\label{figure1}
\end{center}
\end{figure}

Low-rank decomposition is a common method to compress network by matrix or tensor factorization. A series of works \cite{Denton2014Exploiting,Jaderberg2014Speeding,Xue2013Restructuring,Zhang2016Accelerating} have achieved great progress on increasing efficiency of networks by representing the original weights as a form of low-rank. However, these methods couldn't reach an extreme compression ratio with good performance since they may suffer from degeneration problem~\cite{Orhan2017Skip,Saxe2013Exact}, which is harmful for the convergence and performance of the network. Filter group convolution  \cite{Krizhevsky2012ImageNet} is another way to compact the network while keeping independencies among filters, which can alleviate the limitation of degeneration. In this work, we will show a novel method, which decompose a regular convolution to filter group structure~\cite{Ioannou2016Deep} (Fig.~\ref{figure1}), while achieve good performance and large compression ratio at the same time.

The concept of filter group convolution was first used in AlexNet \cite{Krizhevsky2012ImageNet} due to the shortage of GPU's memory. Surprisingly, independent filter groups in CNN learned a separation of responsibility, and the performance was close to that of corresponding network without filter groups, which means this lighter architecture has an equal ability of feature representation. After this work, filter group and depthwise convolution were widely used in designing efficient architectures \cite{Chollet2016Xception,Howard2017MobileNets,Ioannou2016Deep,Sandler2018Inverted,Xie2016Aggregated,Zhang2017ShuffleNet} and achieved state-of-the-art performance among lightweight models. However, all of those well-designed architectures need to be trained from scratch with respect to specific tasks. Huang \emph{et al.}\cite{Huang2017CondenseNet} introduced CondenseNet, which learns filter group convolutions automatically during the training process, while several complicated stages with up to 300 epochs' training are needed to reach both sparsity and regularity of filters.

In this paper, we will show an efficient way to decompose the regular convolutional layer into the form of filter group structure which is a linear combination of filter group convolution and point-wise convolution (Fig.~\ref{figure1}). Taking advantage of filter group convolution, computational complexity of the original network can be extremely compressed while preserve the diversity of feature representation, which leads to faster convergence and less accuracy drop. Besides, our method can be efficiently applied on most of regular pre-trained models without any additional training skill.

The contributions of this paper are summarized as follows:
\begin{itemize}
\item A filter group approximation method is proposed to decompose a regular convolutional layer into a filter group structure with tiny accuracy loss while mostly preserve feature representation diversity.
\item Experiments and discussions provide new inspiration and promising research directions based on degeneracy problem in network compression.
\end{itemize}

\section{Related Work}
\label{Related Work}

First of all, we briefly discuss related works including network pruning, low-rank decomposition and efficient architecture designing.

\textbf{Network pruning:} Network pruning is an efficient method to reduce the redundancy in the deep neural network \cite{Srinivas2015Data}. A straight forward way of pruning is to evaluate the importance of weights (\emph{e.g.} the magnitude of weights \cite{Han2015Learning,Li2016Pruning}, sparsity of activation \cite{Hu2016Network}, Taylor expansion \cite{Molchanov2016Pruning}, \emph{etc.}), thus the less important weights could be pruned with less effect on performance. Yu \emph{et al.}~\cite{Yu2017NISP} proposed a Neuron Importance Score Propagation algorithm to propagate the importance scores to every weight. He \emph{et al.}~\cite{He2017Channel} utilized a Lasso regression based channel selection and least square reconstruction to compress the network. Besides, there are several training-based studies in which the structure of filters are forced to sparse during training \cite{Liu2017Learning,Wen2016Learning}. Recently, Huang \emph{et al.}~\cite{Huang2017CondenseNet} proposed an elaborate scheme to prune neuron connection with both sparsity and regularity during training process.% However, these kind of training-based methods are quite complex and time-consuming.

\textbf{Low-rank decomposition:} Instead of removing neurons of network, a series of works have been proposed to represent the original layer with low-rank approximation. Previous works \cite{Denton2014Exploiting,Lebedev2015Speeding} applied matrix or tensor factorization algorithms (\emph{e.g.} SVD, CP-decomposition) on the weights of each layer to reduce computational complexity. Jaderberg \emph{et al.}~\cite{Jaderberg2014Speeding} proposed a joint reconstruction method to decompose $k\times k$ filter into a combination of $k\times 1$ and $1\times k$ filters. These methods only gain limited compressions on shallow networks. Zhang \emph{et al.}~\cite{Zhang2016Accelerating} improved the low-rank approximation of deeper networks on large dataset using a nonlinear asymmetric reconstruction method. Similarly, Masana \emph{et al.}~\cite{Masana2017Domain} proposed a domain adaptive low-rank decomposition method, which took the activations' statistics of the new domain into account. Alvarez \emph{et al.}~\cite{Alvarez2017} regularized parameter matrix to low-rank during training, such that it could be decomposed easily in the post-processing stage.

\textbf{Efficient architecture designing:} The needs of applying CNNs on resource-constrained devices also encouraged the studies of efficient architecture designing. `Residual' block in ResNet \cite{He2016Deep,He2016Identity}, `Inception' block in GoogLeNet \cite{Szegedy2016Inception,Szegedy2015Going,Szegedy2015Rethinking} and `fire module' in SqueezeNet \cite{Iandola2016Squeezenet} achieved impressive performance with less complexity on spatial extents. AlexNet \cite{Krizhevsky2012ImageNet} utilized filter group convolution to solve the constraints of computational resources. ResNeXt~\cite{Xie2016Aggregated} replaced the regular $3\times 3$ convolutional layers in residual blocks of ResNet with group convolutional layers to achieve better results. Depthwise separable convolution proposed in Xception \cite{Chollet2016Xception} promoted the performance of low-cost CNNs on embedded devices. Benefiting from depthwise separable convolution, MobileNet v1 \cite{Howard2017MobileNets}, v2 \cite{Sandler2018Inverted} and ShuffleNet \cite{Zhang2017ShuffleNet} achieved state-of-the-art performance on several tasks with significantly reduced computational requirements.

\begin{figure}[tb]
\begin{center}
\centerline{\includegraphics[width=0.75\columnwidth]{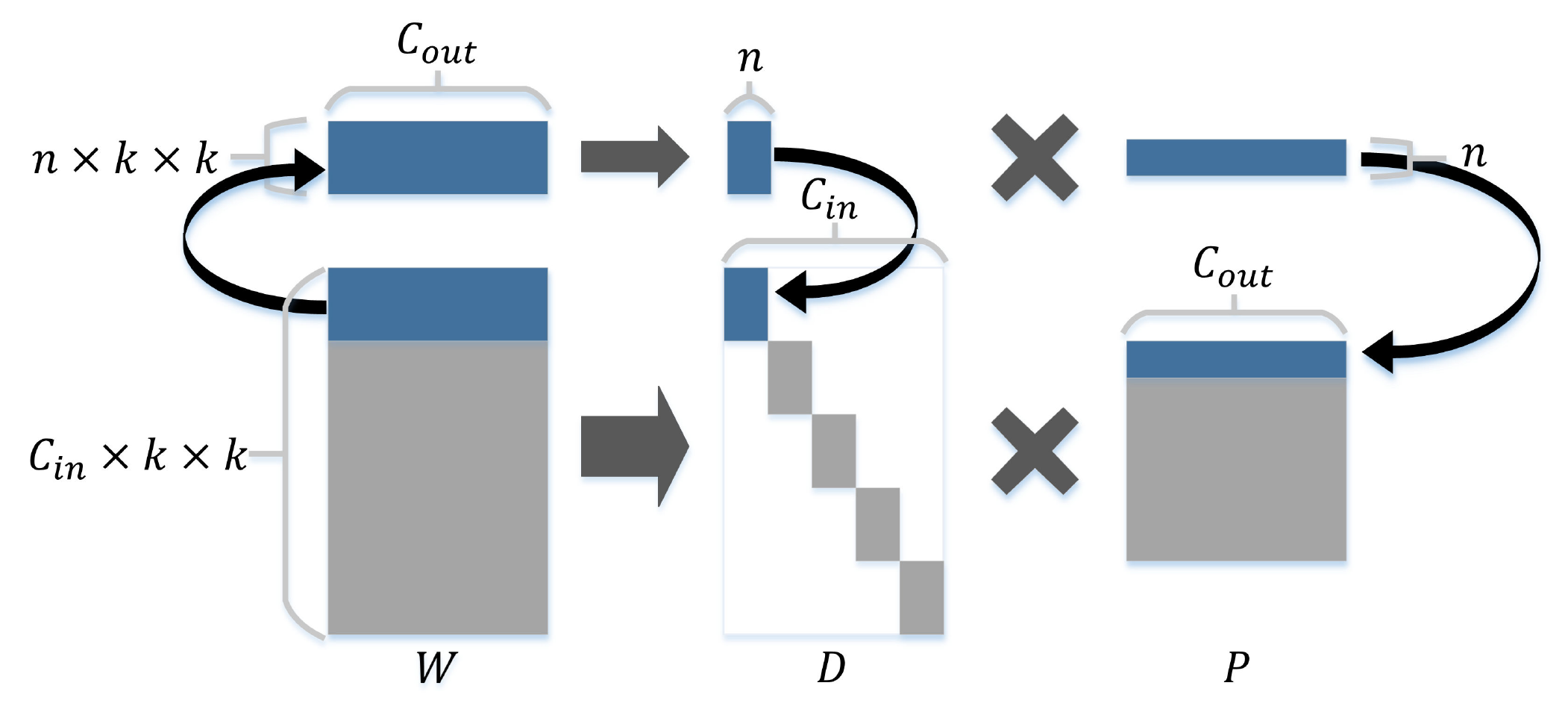}}
\caption{Illustration of the decomposition. The original layer with weight matrix \emph{W} is decomposed into two layer with weight matrices $D$ and $P$ respectively}
\label{figure2}
\end{center}
\end{figure}

\section{Approaches}
\label{Approaches}

In this section we introduce a novel filter group approximation method to decompose a regular convolutional layer into the form of filter group structure. Furthermore, we discuss the degeneracy problem of the compressed network.

\subsection{Filter Group Approximation of Weights}
\label{method}

Weights of convolutional layer can be considered as tensor $W \in \mathbb{R}^{C_{in}\times k\times k\times C_{out}}$, where $C_{out}$ and $C_{in}$ are the number of output channels and input channels respectively, and $k$ is the spatial size of filters. The response $Y \in \mathbb{R}^{N\times C_{out}}$ is computed by applying $W$ on $X \in \mathbb{R}^{N\times C_{in}\times k\times k}$ which is sampled from $k \times k$ sliding window of the layer inputs. Thus the convolutional operation can be formulated as:
\begin{equation}
\label{eq1}
    Y = X\times W.
\end{equation}
The bias term is omitted for simplicity. $X$ and $W$ can be seen as matrices with shape $N$-by-($C_{in}k^{2}$) and ($C_{in}k^{2}$)-by-$C_{out}$ respectively. The computational complexity of Eqn.~\ref{eq1} is $O(C_{in} k^2 C_{out} )$.

Each row of matrix $W$ is only multiplied with the corresponding column of matrix $X$. Let's consider each sub-matrix with $n \times k\times k$ rows  of $W$ and $n \times k\times k$ columns of $X$ ($n$ is divisible by $C_{in}$), which are denoted as $W_i$ and $X_i$ ($i=1,2,\ldots,\left.C_{in}\middle/n\right.$) respectively. Eqn.~\ref{eq1} can be equally described as:
\begin{equation}
\label{eq2}
    Y = \begin{bmatrix} X_1 & X_2 & ... & X_{\left.C_{in}\middle/n\right.}  \end{bmatrix}\begin{bmatrix} W_1\\ W_2\\ ...\\ W_{\left.C_{in}\middle/n\right.} \end{bmatrix} .
\end{equation}

Each sub-matrix $W_i$ with rank $d_i$ can be decomposed into two matrices $D_i$ and $P_i$ using SVD. Considering the redundancy of parameter in a layer, $W_i$ can be approximated by $ \widetilde{W_i} = D_{i,n}\times P_{i,n}^T$ with rank $n \leq d_i$. $D_{i,n}$ and $P_{i,n}$ are the first $n$ columns of $D_i$ and $P_i$ related to the largest $n$ singular values. By arranging $D_{i,n}$ and $P_{i,n}$ from each sub-matrices of $W$ into two matrices denoted as $D$ and $P$ respectively, we can get an approximation of matrix $W$ which is $W \approx D\times P$ (Fig.~\ref{figure2}). The original layer with weights $W$ can be replaced by two layers with weights $D$ and $P$ respectively. Thus the original response can be approximated by $Y^{*}$ with:
\begin{equation}
\label{eq3}
    Y \approx Y^{*} = X\times D\times P.
\end{equation}

\setlength{\tabcolsep}{2pt}
\begin{table}[t]
\begin{center}
\caption{Settings of \emph{n} for each stage in whole network compression on ResNet34. `\emph{Ours-Res34/A}', `\emph{Ours-Res34/B}', `\emph{Ours-Res34/C}' and `\emph{Ours-Res34/D}' denote four compression networks using our method. `-' means no compression in this stage}
\label{arch}
\begin{tabular}{c|c|c|c|c|c|c}
\hline
{\scriptsize layer name} & \scriptsize output size & \scriptsize ResNet34  & \scriptsize Ours-Res34/A & \scriptsize Ours-Res34/B & \scriptsize Ours-Res34/C & \scriptsize Ours-Res34/D\\
\hline
\scriptsize conv1 & \scriptsize $112\times 112$ & \multicolumn{5}{c}{-}\\
\hline
{\scriptsize conv2\_x} & \scriptsize $56\times 56$ & - & \scriptsize $n=8$ & \scriptsize $n=4$ &  \scriptsize $n=1$ &  \scriptsize $n=1$\\

\hline
{\scriptsize conv3\_x} &  \scriptsize $28\times 28$  & -  & \scriptsize $n=32$  &  \scriptsize $n=16$   & \scriptsize  $n=4$ & \scriptsize $n=1$\\

\hline
{\scriptsize conv4\_x} & \scriptsize $14\times 14$  & -  &\scriptsize $n=128$  & \scriptsize $n=64$  &\scriptsize $n=16$ & \scriptsize $n=1$\\
\hline
{\scriptsize conv5\_x} & \scriptsize $7\times 7$  & -  & - &\scriptsize $n=256$  & \scriptsize $n=64$ & \scriptsize $n=1$\\
\hline
{\scriptsize fc}& \scriptsize $1\times 1$  & \multicolumn{5}{c}{-} \\
\hline
\multicolumn{2}{c|}{\scriptsize FLOPs} & \scriptsize $7.32\times10^9$  &\scriptsize $3.98\times10^9$  &\scriptsize $2.58\times10^9$  &\scriptsize $1.44\times10^9$  &  \scriptsize $1.11\times10^9$\\
\hline
\end{tabular}
\end{center}
\end{table}

The matrix $D$ is a block diagonal matrix as illustrated in Fig.~\ref{figure2}, which can be implemented using a group convolutional layer with $C_{in}$ filters of spatial size $k\times k$, and each filter convolve with $n$ input channels sequentially. $P$ is a $1\times 1$ convolutional layer to create a linear combination of the output of $D$. So the computational complexity of Eqn.~\ref{eq3} is $O(C_{in} k^2 n + C_{in} C_{out} )$. Compare with Eqn.~\ref{eq1}, the complexity reduces to $(\left.n\middle/C_{out}\right.+\left.1\middle/k^2\right.)$. It shrinks to about $\left.1\middle/k^2\right.$ of the original one in the case of $n=1$, which is known as depthwise convolution.

\subsection{Reconstruction and Fine-tuning for the Compressed Network}
Since our approximation is based on sub-matrices of $W$, the accumulative error and magnitude difference among approximated sub-matrices will damage the overall performance. Hence we further minimize the reconstruction error by:
\begin{equation}
\label{eq4}
    A = arg\min_{A^*}\parallel{Y}-Y^{*}\times A^{*}\parallel_2^2,
\end{equation}
where ${Y}$ indicates the response of the original network, and $Y^*$ is the response after approximation. Eqn.~\ref{eq4} is a typical linear regression problem without any constrains which can be solved by the least-square optimization. Matrix $A$ can be implemented as a $1\times 1$ convolutional layer, which can be merged into $P$ in practice, thus there is no additional layer.

After reconstruction for each layer, the compressed network can maintain good performance even without fine-tuning. In our experiments, few epochs' fine-tuning (usually less than 20 epochs) with a very small learning rate is enough to achieve better accuracy. %Fig.~\ref{figure1} illustrates the filter group structure decomposed from a convolutional layer using our proposed method.

\begin{figure}[t]
\centering
\subfigure[]{
\begin{minipage}{5.8cm}
\centering
\includegraphics[width=0.95\columnwidth]{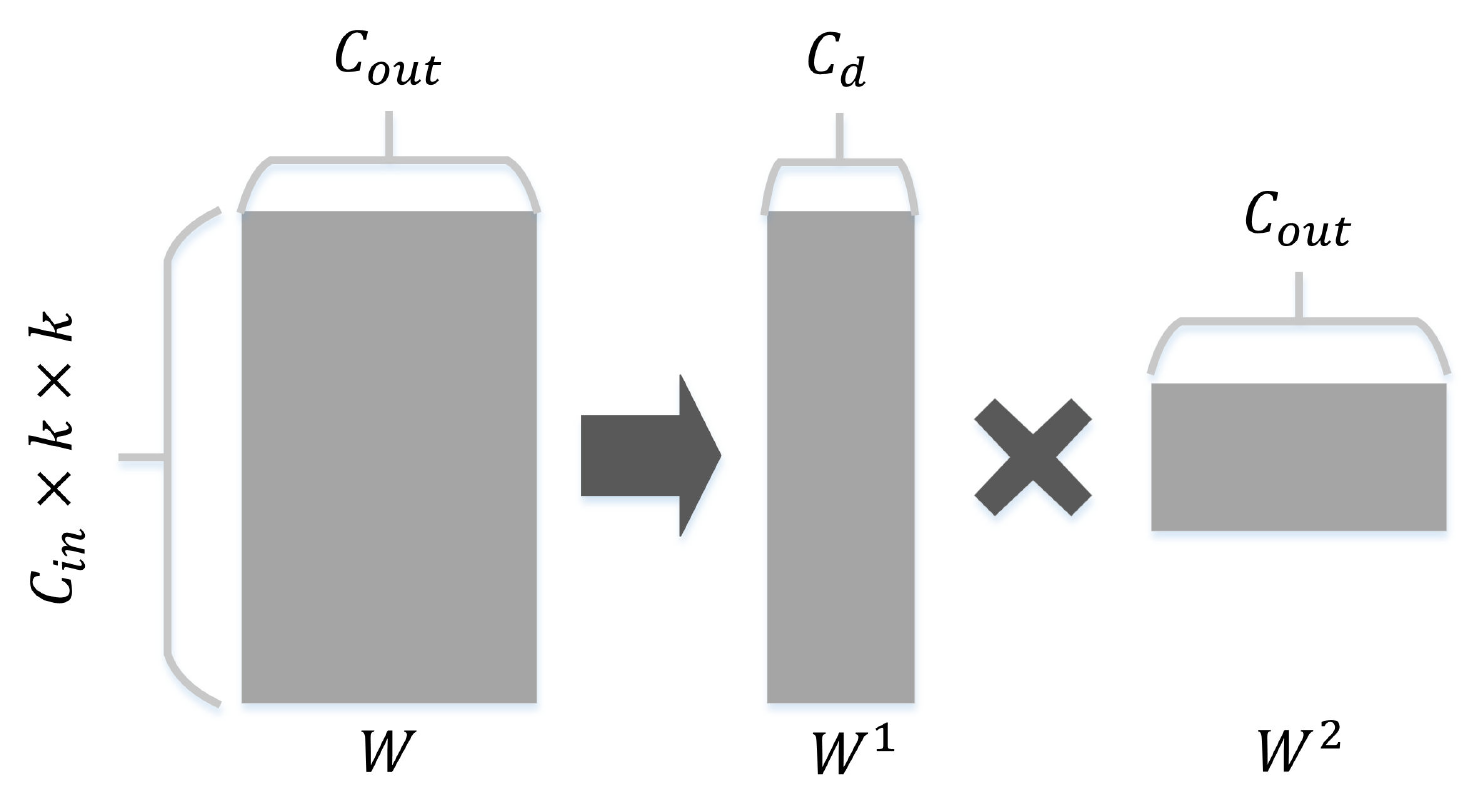}
\label{fig:subfig:b}
\end{minipage}
}
\subfigure[]{
\begin{minipage}{5.8cm}
\centering
\includegraphics[width=0.97\columnwidth]{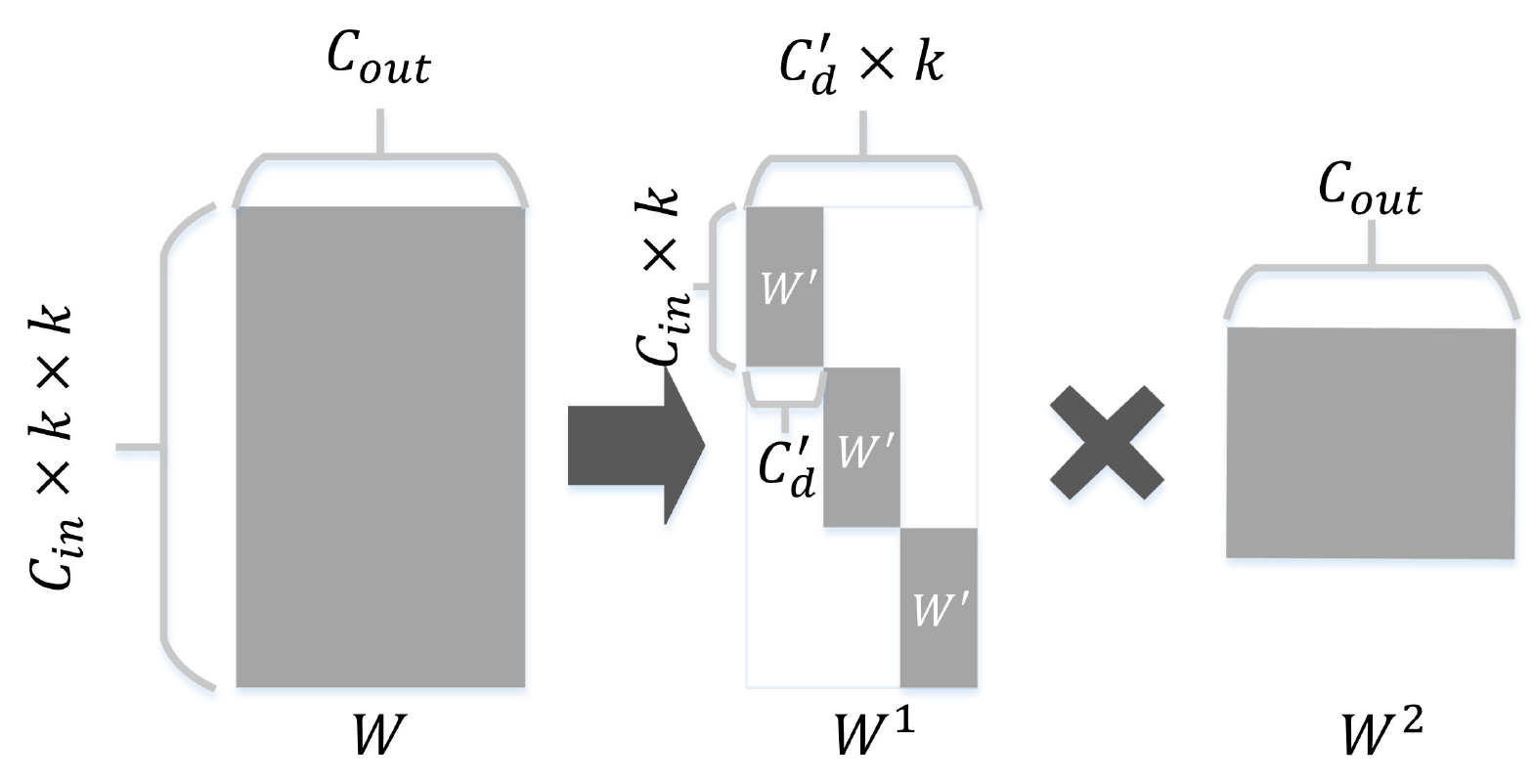}
\label{fig:subfig:c}
\end{minipage}
}
\caption{Two low-rank decomposition strategies. (a) The SVD based method. The first $C_{d}$
singular vectors are reserved in SVD. (b) The reconstruction method using $k\times 1$ and $1\times k$ convolution. The block diagonal weight matrix $W^{1}$ is constructed using $k$ replicas of $W^{'}$, which is weight matrix of the $k\times 1$ convolution. $W^{2}$ is weight matrix of the $1\times k$ convolution } %                         % 大图名称
\label{figure3}                                                        %图片引用标记
\end{figure}

\subsection{Compression Degree for Each Layer}
\label{degree}
When compressing an entire network, a proper compression ratio need to be determined for each layer. One common strategy is removing the same proportion of parameters for each layer. However, it is not reasonable since different layers are not equally redundant. As mentioned by \cite{He2017Channel,Zhang2016Accelerating}, deeper layers have less redundancy, which indicates less compression with increasing depth in whole network compression.

In our method, the compression ratio is controlled by \emph{n} (see Section~\ref{method}). For a whole network compression, we set larger \emph{n} for deeper layers. Convolutional layers of the whole network can be separated into several stages according to the spatial size of corresponding output feature. Taking the experience of \cite{Ioannou2016Deep} for reference, we set the same \emph{n} for those layers in the same stage, and the ratio of two values of \emph{n} between adjacent stages is $1:4$ (shallow one \emph{vs} deep one). Table~\ref{arch} shows settings of \emph{n} for each stage of ResNet34 in whole network compression, and `\emph{Ours-Res34/D}' is a `depthwise' compressed case in which \emph{n} for all stages is 1. We will further discuss different degrees of compression with network depth in our experiment.

\begin{figure}[tb]
\centering
\subfigure[\scriptsize VGG16(S) on CIFAR100]{
\begin{minipage}{3.7cm}
\centering
\includegraphics[scale=0.25]{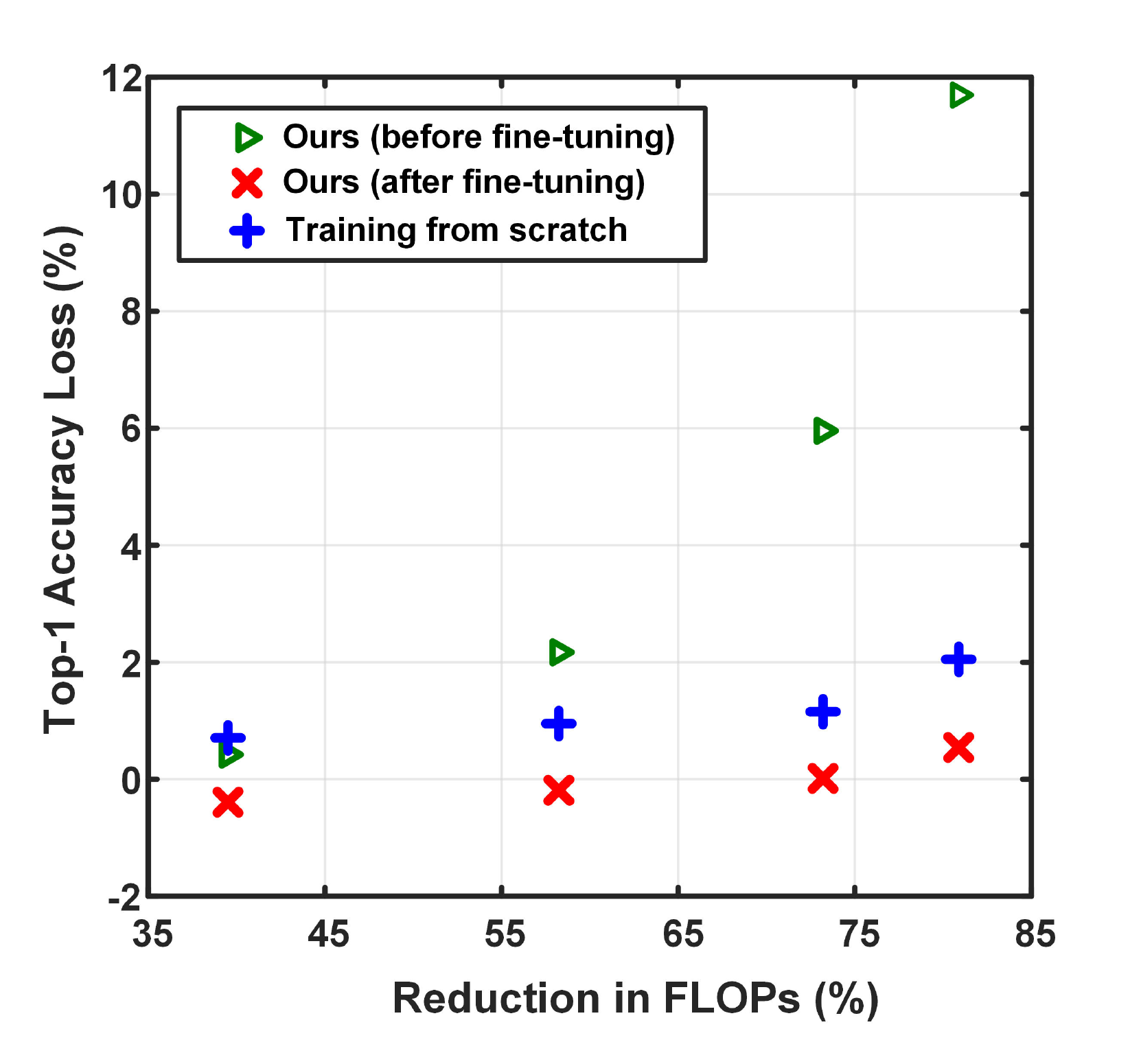}
\end{minipage}
}
\subfigure[\scriptsize VGG16 on ImageNet]{
\begin{minipage}{3.65cm}
\centering
\includegraphics[scale=0.25]{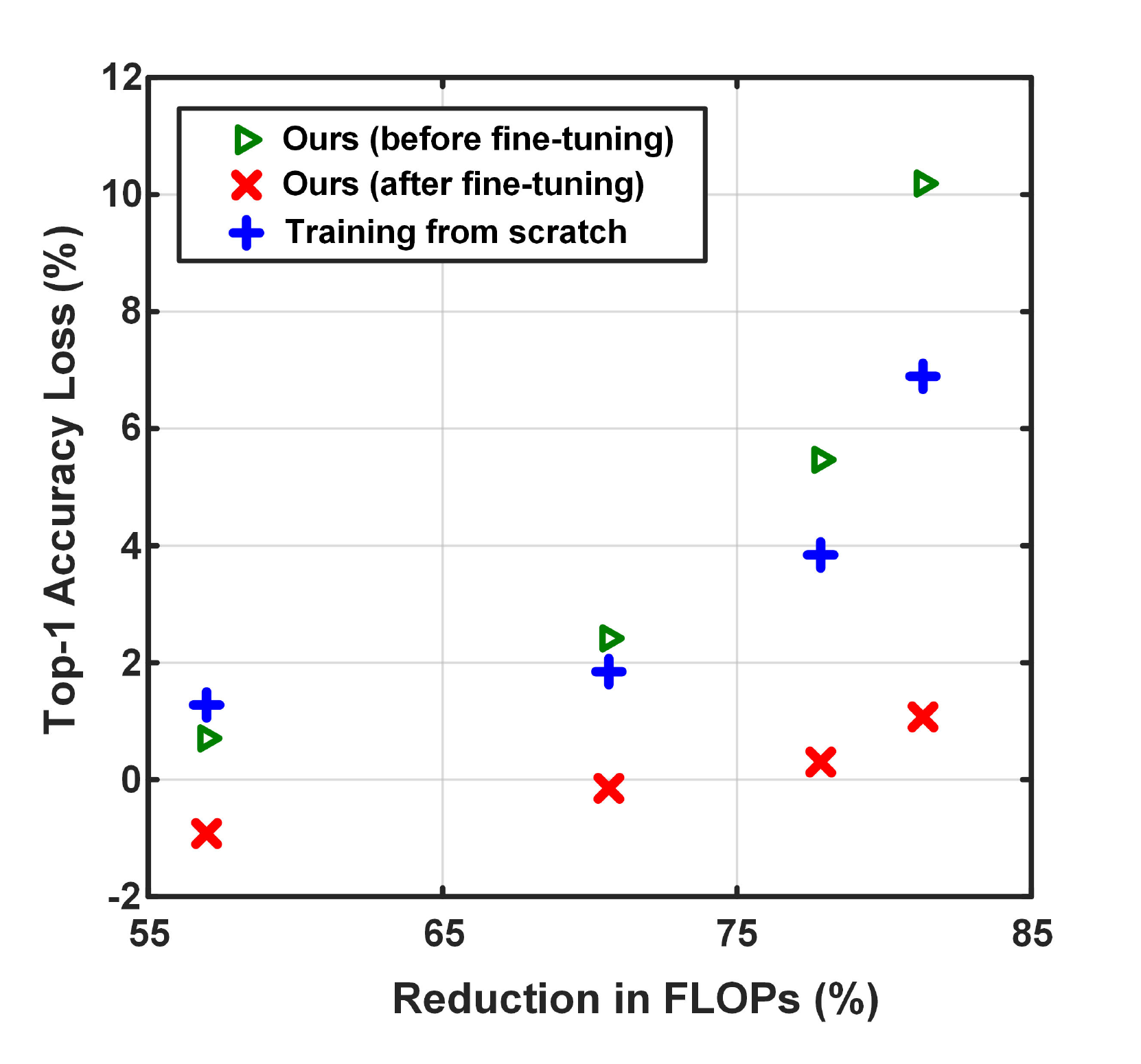}
\end{minipage}
}
\subfigure[\scriptsize ResNet34 on ImageNet]{
\begin{minipage}{4cm}
\centering
\includegraphics[scale=0.25]{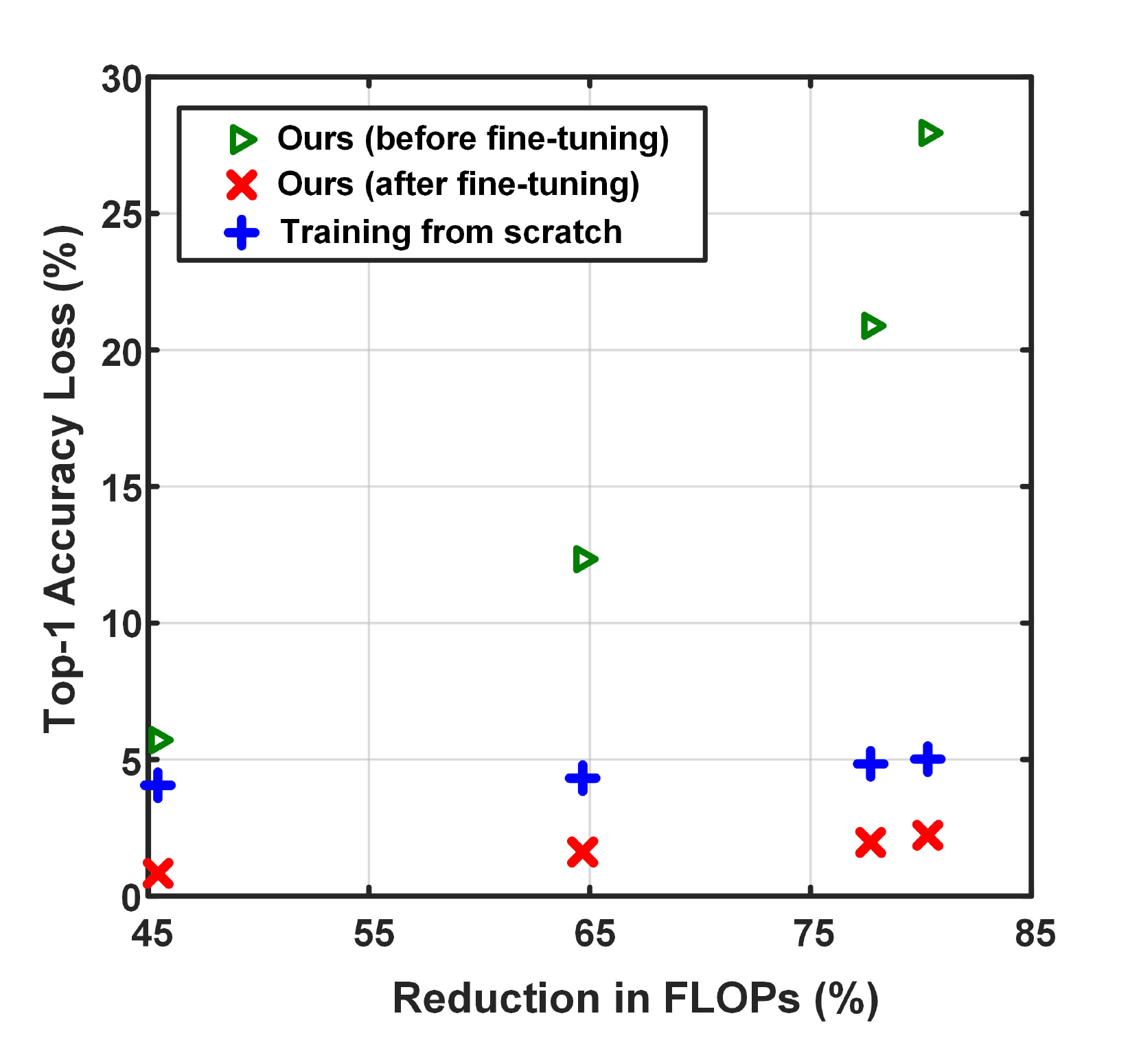}
\end{minipage}
}
\caption{Comparison with training from scratch. The green `$\triangleright$' denotes the top-1 accuracy loss of compressed network without fine-tuning; the red `$\times$' denotes the top-1 accuracy loss after fine-tuning; the blue `$+$' denotes the top-1 accuracy loss of training from scratch. A negative value of y-axis indicates an improved model accuracy} %                         %大图名称
\label{figure4}                                                        %图片引用标记
\end{figure}

\subsection{Consideration about the Degeneracy Problem}
\label{degeneracy}

As mentioned in \cite{Saxe2013Exact}, the Jacobian of weights indicates the correlation between inputs and outputs, and degeneracy of Jacobian leads to poor backpropagation of gradients which impacts the convergence and performance of the network. Jacobian can be computed as: $J=\left.\partial Y\middle/\partial X\right.$, where $X$ and $Y$ are inputs and outputs respectively.
For linear case of two layers with weights $W^{1}$ and $W^{2}$, the Jacobian $J = W^{1}\times W^{2}$.

Low-rank decomposition method achieves compression by representing the original layer with a linear combination of several layers on a low-rank subspace. Fig.~\ref{figure3} illustrates two common low-rank decomposition strategies that decompose the weight matrix $W$ into two matrices $W^{1}$ and $W^{2}$. $C_{in}$ and $C_{out}$ are numbers of input and output channels respectively, $k$ is the spatial size of filters. Let's consider the case that $C_{in} \leq C_{out} < C_{in}k^{2}$ $(k>1)$, which is common in regular convolutional layers of classic networks like VGG~\cite{Simonyan2014Very}, ResNet~\cite{He2016Deep}.

Fig.~\ref{fig:subfig:b} shows an instance of SVD based method \cite{Zhang2016Accelerating,Masana2017Domain}. The first $C_d$ singular vectors are reserved in SVD. Thus the rank of Jacobian is $R_{1} = C_{d}$,\footnote{We suppose that weight matrices after decomposition are full-rank which always holds in practice.} and the computational complexity of the low-rank representation is $O(C_{in} k^2 C_{d} + C_{d} C_{out} )$. While in our method (Fig.~\ref{figure2}), the rank $R_{ours}$ can reach to $C_{in}$ no matter how much the model is compressed, and the computational complexity is $O(C_{in} k^2 n + C_{in} C_{out} )$. Under the same compression ratio, we can get:
\begin{equation}
\label{eq5}
    C_{d} = \frac{{C_{in} k^2 n + C_{in} C_{out}}}{{C_{in} k^2 + C_{out}}}.
\end{equation}
Since $n<C_{in}$, $R_{1} <R_{ours}$. Eqn.~\ref{eq5} shows that, the more the model is compressed, the less $C_d$ is, making the Jacobian more degenerate.

Jaderberg \emph{et al.}~\cite{Jaderberg2014Speeding} proposed a joint reconstruction method to represent a $k\times k$ filter with two $k\times 1$ and $1\times k $ filters, which is equivalent to the representation of Fig.~\ref{fig:subfig:c}. $W^{'}$ is the weight matrix of the $k\times 1$ convolutional layer with $C_{d}^{'}$ output channels. The rank of Jacobian is $R_{2} = \min(C_{d}^{'}k, C_{out}) $, and the computational complexity is $O(C_{in} C_{d}^{'} k + C_{d}^{'} k C_{out} )$. Similarly, under the same compression ratio,
\begin{equation}
\label{eq5}
    C_{d}^{'} k = \frac{{C_{in} k^2 n + C_{in} C_{out}}}{{C_{in} + C_{out}}}.
\end{equation}
When $n < \left.C_{in}\middle/k^2\right.$, $R_{2} <R_{ours}$, which means the compressed network from~\cite{Jaderberg2014Speeding} degrades more quickly than ours as the compression ratio increases.

%Low-rank decomposition method achieves compression by representing the original layer with a linear combination of few layers on a low-rank subspace. Fig.~\ref{fig:subfig:c} illustrates a commonly used low-rank decomposition strategy \cite{Denton2014Exploiting,Zhang2016Accelerating,Masana2017Domain}, the weight matrix $W$ is decomposed into two matrices $W^{12}$ and $W^{23}$ using SVD, the first $C_d$ singular vectors are reserved in decomposition (consider the situation that $C_{out}<C_{in} \times k \times k$). Thus the rank of Jacobian is $C_d < C_{out}$. The more the model is compressed, the less $C_d$ is, making the Jacobian degenerate. Other low-rank decomposition methods also suffer from similar problems. While for our method (Fig.~\ref{figure2}), the rank of Jacobian can reach to $\min(C_{in}, C_{out})$ theoretically no matter how much the model compressed, thus the degeneracy of the compressed network can be efficiently alleviated. We will further prove it in our experiments.

\section{Experiments}

In this section, a set of experiments are performed on standard datasets with commonly used CNN networks to evaluate our method. Our implementations are based on Caffe \cite{Jia2014Caffe}.
%We compare our method with training from scratch and several recently proposed methods to demonstrate the effectiveness of our method. We mainly focus on the reduction in the number of FLOPs. Accuracy losses are reported in our experiments. Our implementations are based on Caffe \cite{Jia2014Caffe}.

\begin{table}[tb]
\begin{center}
\caption{Comparison with existing method for VGG16(S) on CIFAR100. FLOPs$\downarrow$ denotes the compression of computations; Top-1$\downarrow$ denotes the top-1 accuracy loss, the lower the better. A negative value here indicates an improved model accuracy}
\label{table1}
\begin{tabular}{lrr}
\toprule
     Method               & FLOPs$\downarrow$ & Top-1$\downarrow$ \\
%\noalign{\smallskip}
\midrule
{Ours-VGG16(S)/A}                     & \textbf{40.03\%} & \textbf{-0.39\%} \\
{\emph{Asym.} {\cite{Zhang2016Accelerating}} (\emph{Fine-tuned, our impl.})    }     & {39.77\%} & {0.28\%}\\
%{Li \emph{et al.} {\cite{Li2016Pruning}} (\emph{Our impl.})}     & 38.51\% & 0.53\% \\
{Liu \emph{et al.} {\cite{Liu2017Learning}}}  & 37.00\% & -0.22\% \\
\midrule
{Ours-VGG16(S)/B}                      & \textbf{73.40\%} & \textbf{0.02\%} \\
{\emph{Asym.} (\emph{Fine-tuned, our impl.})    }     & {73.21\%} & {1.03\%}\\
%{Li \emph{et al.} {\cite{Li2016Pruning}} (\emph{Our impl.})}     & 73.15\% & 2.92\% \\
{Liu \emph{et al.} }  & 67.30\% & 2.34\% \\
\midrule
{Ours-VGG16(S)/C}                       & \textbf{84.82\%} & \textbf{0.57\%} \\
{\emph{Asym.} (\emph{Fine-tuned, our impl.})    }     & {84.69\%} & {2.94\%}\\
%{Li \emph{et al.} {\cite{Li2016Pruning}} (\emph{Our impl.})}     & 84.57\% & 4.64\% \\
{Liu \emph{et al.} }  & 83.00\% & 3.85\% \\
\midrule
{Ours-VGG16(S)/D}                       & \textbf{88.58\%} & \textbf{1.93\%} \\
{\emph{Asym.}  (\emph{Fine-tuned, our impl.})    }     & {87.33\%} & {4.47\%}\\
{Liu \emph{et al.}  \footnotemark[2]}  & - & - \\
\bottomrule
\end{tabular}
\end{center}
\end{table}
\footnotetext[2]{Liu \emph{et al} has not provided result with compression ratio about 88\% .}

\setlength{\tabcolsep}{4pt}
\begin{table}[!htb]
\begin{center}
\caption{Comparison with existing method for VGG16 and ResNet34 on ImageNet. FLOPs$\downarrow$ denotes the compression of computations; Top-1$\downarrow$ and Top-5$\downarrow$ denote the top-1 and top-5 accuracy losses, the lower the better}
\label{table2}
\begin{tabular}{llrrr}
\toprule
 Network           &Method        & FLOPs$\downarrow$ & Top-1$\downarrow$ & Top-5$\downarrow$\\ \midrule
%\noalign{\smallskip}
\multirow{2}{*}{VGG16}  &{Ours-VGG16/A}                      & \textbf{56.99\%} & \textbf{-0.82\%} & \textbf{-0.94\%} \\
                        &{\emph{Asym.} (\emph{Fine-tuned, our impl.})    }     & {56.17\%} & {-0.18\%} & {-0.27\%}\\
                        &{He \emph{et al.} {\cite{He2017Channel}}   }     & {$\approx$ 50\%} & {-} & {0.0\%} \\
\cmidrule{2-5}        &{Ours-VGG16/B}                      & \textbf{77.86\%} & {0.28\%}  & {0.07\%}\\
                        &{\emph{Asym.}  (\emph{Fine-tuned, our impl.})    }     & {77.60\%} & {1.81\%} & {0.73\%}\\
&{\emph{Asym.} (3D)  {\cite{Zhang2016Accelerating}}  }     & {$\approx$ 75\%} & {-} & {0.3\%} \\
&{He \emph{et al.}    }     & {$\approx$ 75\%} & {-} & {1.0\%} \\
&{He \emph{et al.} (3C) {\cite{He2017Channel}}   }     & {$\approx$ 75\%} & {-} & \textbf{0.0\%} \\
&{Jaderberg \emph{et al.} {\cite{Jaderberg2014Speeding}} (\cite{Zhang2016Accelerating}\emph{'s impl.})  }     & {$\approx$ 75\%} & {-} & {9.7\%} \\
\cmidrule{2-5}
&{Ours-VGG16/C}                       & \textbf{81.35\%} & \textbf{1.06\%} & \textbf{0.27\%}\\
&{\emph{Asym.} (\emph{Fine-tuned, our impl.})    }     & {81.18\%} & {4.53\%} & {2.45\%}\\
&{\emph{Asym.} (3D)   }     & {$\approx$ 80\%} & {-} & {1.0\%} \\
&{He \emph{et al.}  }     & {$\approx$ 80\%} & {-} & {1.7\%} \\
&{He \emph{et al.} (3C)  }     & {$\approx$ 80\%} & {-} & {0.3\%} \\
&{Kim \emph{et al.} {\cite{Kim2015Compression}}   }     & { 79.72\%} & {-} & {0.5\%} \\
\cmidrule{2-5}
&{Ours-VGG16/D}                       & \textbf{85.80\%} & \textbf{3.49\%} &\textbf{2.03\%} \\
&{\emph{Asym.} (\emph{Fine-tuned, our impl.})  }     & {83.91\%} & {6.96\%} & {5.14\%} \\
&{\emph{Asym.} (3D)  (\emph{Fine-tuned, our impl.}) }     & {84.44\%} & {4.51\%} & {2.89\%} \\
&{He \emph{et al.} (3C)   }     & { 85.55\%} & {4.38\%} & {2.96\%} \\
\midrule
\multirow{2}{*}{ResNet34 }&{Ours-Res34/A}                      & \textbf{45.63\%} & \textbf{0.35\%} & \textbf{0.04\%}  \\
&{Li \emph{et al.}~\cite{Li2016Pruning} }     & {24.20\%} & {1.06\%} & {-} \\
&{\emph{NISP-34-B} {\cite{Yu2017NISP}} }     & {43.76\%} & {0.92\%} &{-} \\
&{\emph{Asym.}  (\emph{Fine-tuned, our impl.}) }     & {44.60\%} & {0.97\%} &{0.21\%} \\
\cmidrule{2-5}
&{Ours-Res34/B}                      & \textbf{64.75\%} & \textbf{1.02\%} & \textbf{0.30\%}\\
&{Li \emph{et al.} (\emph{Fine-tuned, our impl.}) }     & {63.98\%} & {4.35\%} &{2.29\%} \\
&{\emph{Asym.}  (\emph{Fine-tuned, our impl.}) }     & {64.12\%} & {2.91\%} &{1.41\%} \\
%{\emph{Asym.} (3D)  {\cite{Zhang2016Accelerating}} (\emph{Fine-tuned, our impl.}) }     & {$\approx$ 64.52\%} & {-} & {0.3\%} \\
\cmidrule{2-5}
&{Ours-Res34/C}                       & \textbf{80.33\%} & \textbf{1.70\%} & \textbf{0.44\%}\\
&{Li \emph{et al.} (\emph{Fine-tuned, our impl.}) }   & {79.67\%} & {8.23\%} &{4.96\%} \\
&{\emph{Asym.}  (\emph{Fine-tuned, our impl.}) }     & {79.95\%} & {6.77\%} &{4.10\%} \\
&{\emph{Asym.} (3D)  (\emph{Fine-tuned, our impl.}) }     & {80.08\%} & {3.79\%} & {1.85\%} \\
\cmidrule{2-5}
&{Ours-Res34/D}                       & 84.84\% & \textbf{3.03\%} & \textbf{1.22\%}\\
&{Li \emph{et al.}  (\emph{Fine-tuned, our impl.}) }    & \textbf{84.99\%} & {13.27\%} &{8.84\%} \\
&{\emph{Asym.} (\emph{Fine-tuned, our impl.}) }     & {84.79\%} & {12.30\%} &{8.27\%} \\
&{\emph{Asym.} (3D) (\emph{Fine-tuned, our impl.}) }     & {84.04\%} & {5.35\%} & {2.78\%} \\
%\midrule
%\multirow{2}{*}{ResNet50 }&{Ours-Res50/A}                      & \textbf{50.12\%} & \textbf{0.78\%} & \textbf{0.40\%}  \\
%&{He \emph{et al.} (3C) {\cite{He2017Channel}}   }     & {$\approx$ 50\%} & {-} & {1.4\%} \\
\bottomrule
\end{tabular}
\end{center}
\end{table}

\begin{table}[t]
\begin{center}
\caption{Comparison of actual running time (ms) per image (224$\times$224). }
\label{times}
\begin{tabular}{llccccc}
\toprule
{ Network} &  Methods &  FLOPs$\downarrow$  &  Top-5$\downarrow$ &  GPU &  CPU &  TX1\\
\hline
\multirow{2}{*}{VGG16 } &  baseline & {-}      & {-}     &  2.91  &  2078 &  122.36\\
                  &  Ours     &  \textbf{85.80\%} &  \textbf{2.03\%} &  2.17  &   \textbf{628} &   \textbf{40.92}\\
                  &  \emph{Asym.}(3D)&  84.44\% &  2.89\% &  1.78  &   778 &   97.36\\
                  &  He \emph{et al.}(3C)&  85.55\% &  2.96\% &  1.65  &   741 &   94.27\\
\midrule
\multirow{2}{*}{ResNet34 } &  baseline & {-}      & {-}     &  0.75  &  442 &  20.5\\
                  &  Ours     &  \textbf{84.84\%} &  \textbf{1.22\%} &  \textbf{0.62}  &   \textbf{116} &   \textbf{9.54}\\
                  &  \emph{Asym.}(3D)&  84.04\% &  2.78\% &  0.67 &    128 &   18.33\\
\bottomrule
\end{tabular}
\end{center}
\end{table}

\subsection{Datasets and Experimental Setting}

%We evaluate the performance of compression on VGG16~\cite{Simonyan2014Very} and ResNet34~\cite{He2016Deep}.
Focusing on the reduction of FLOPs and accuracy drops, we compare our method with training from scratch and several state-of-the-art methods on CIFAR100 and ImageNet-2012 datasets~\cite{Deng2009ImageNet} for image classification task.

CIFAR100 dataset contains 50,000 training samples and 10,000 test samples with resolution $32\times 32$ from 100 categories. We use a standard data augmentation scheme~\cite{He2016Deep} including shifting and horizontal flipping in training. A variation of the original VGG16 from~\cite{Liu2017Learning} (called VGG16(S) in our experiments) with top-1 accuracy 73.26\% is used to evaluate the compression on CIFAR100 dataset.

ImageNet dataset consists of 1.2 million training samples and 50.000 validation samples from 1000 categories. We resize the input samples with short-side 256 and then adopt $224\times 224$ crop (random crop in training and center crop in testing). Random horizontal flipping is also used in training. We evaluate single-view validation accuracy. On ImageNet dataset, we adopt the compression on VGG16~\cite{Simonyan2014Very} and ResNet34~\cite{He2016Deep}. %The top-1/top-5 accuracy of VGG16 on ImageNet is 70.97\%/89.99\%. ResNet34 is a more efficiently designed architecture, which makes the compression more challenging. The top-1/top-5 accuracy of ResNet34 is 73.27\%/91.24\%.

\subsection{Comparison with Training-from-scratch and Existing Methods}

\subsubsection{Comparison with Training-from-scratch}
Several compression ratios are adopted in our experiments to evaluate the performance. For the fine-tuning processes of the compressed networks, we use an initial learning rate of 1e-4, and divide it by 10 when training losses are stable. $5 \sim 20$ epochs are enough for fine-tuning to achieve convergence. For comparison, we directly train models with the same architecture as our compressed networks from scratch for more than 100 epochs, with the same initialization and learning strategies as proposed by the original works~\cite{He2016Deep,Simonyan2014Very}.

%For comparison, we directly train the same architecture as our compressed networks from scratch. We set the same initialization and learning strategies as those of original networks. The processes of training from scratch take more than 100 epochs.

Comparison results are shown in Fig.~\ref{figure4}. With 57\% reduction of FLOPs for VGG16, the compressed network without fine-tuning only has a tiny accuracy loss on ImageNet, and it's comparable to the performance of the scratch counterpart. With a few epochs' fine-tuning, the compressed networks outperform training from scratch by a large margin. In some cases, our method even achieves higher accuracy than the original network.

\subsubsection{Comparison with Existing Methods}

Next, we compare our method with several state-of-the-art methods under the same compression ratios. To be noted, results demonstrated in our experiments are all  fine-tuned.
%However, there are only limited results of these related works, thus some of these comparisons are implemented by us. We demonstrate the results with fine-tuning in our experiments.

As shown in Table~\ref{table1}, for VGG16(S) on CIFAR100, 4 compressed architectures (\emph{`Ours-VGG16(S)/A,B,C,D'}) are created by using different \emph{n} as mentioned in Section~\ref{degree}. \emph{`Ours-VGG16(S)/A'} increases top-1 accuracy by 0.39\% with 40\% FLOPs reduction, which is slightly better than the other two methods. Under larger compression ratios, our method outperforms the other two methods with much lower accuracy loss. With 88.58\% reduction of FLOPs (8.5$\times$ speed-up), the increased error rate of \emph{`Ours-VGG16(S)/D'} is 1.93\% while \emph{Asym.}'s is 4.47\% under a similar compression ratio.

Networks trained on Imagenet with complex features are less redundant, which make compressing such networks much more challenging. The results of VGG16, ResNet34 and ResNet50 on ImageNet are shown in Table~\ref{table2}.

For VGG16, we adopt architectures \emph{`Ours-VGG16/A,B,C,D'} with respect to 4 compression ratios. With 56.99\% reduction of FLOPs, the top-5 accuracy of \emph{`Ours-VGG16/A'} is 0.94\% higher than the original network. When we speed the network up to $7\times$, \emph{`Ours-VGG16/D'} can still achieve result of 2.03\% top-5 accuracy drop, while for \emph{Asym.}~\cite{Zhang2016Accelerating}, it is much larger (5.14\% drop on top-5 accuracy). \emph{Asym.} (3D) and He \emph{et al.} (3C)~\cite{He2017Channel} improve their accuracy under the same compression ratios by combining with spatial decomposition \cite{Jaderberg2014Speeding}. However, the compressed architectures from combined methods are much inefficient on most of real-time devices since the original convolutional layer is decomposed into three layers with spatial filter size $3\times1$, $1\times3$ and $1\times1$. Besides, the combination can be also applied on our method. We will conduct the exploration in our future work.

For ResNet34, the advantages in previous experiments still hold. \emph{`Ours-Res34/A'} reduces 45.63\% of FLOPs with negligible accuracy drop. As the compression ratio increases, the performance of our method degrades very slowly. \emph{`Ours-Res34/D'} achieves 1.22\% top-5 accuracy loss with 84.84\% FLOPs reduction ($6.6\times$ speed-up), while Li \emph{et al.}~\cite{Li2016Pruning} and \emph{Asym.} suffer rapidly increased error rate. Similar to the results of VGG16, the combined \emph{Asym.} (3D) outperforms \emph{Asym.} a lot, but it is still worse than ours.

%For ResNet50, since $1\times 1$ layer can hardly be decomposed, which is one of our future works, our method can only achieved $1.6\times$ compression.
\subsection{Actual Speed up of the Compressed Network}
We further evaluate the actual running time of the compressed network. Table~\ref{times} illustrates time cost comparisons on different devices including Nvidia TITAN-XP GPU, Xeon E5-2650 CPU and Nvidia Jetson TX1. Due to the inefficient implementation of group operation in Caffe-GPU, our method shows large time cost. However, in most of resource-constrained devices, like CPU and TX1, bandwidth becomes the main bottleneck. In such case, our method achieves much significant actual speed up as shown in Table~\ref{times}.

\begin{figure}[t]
\begin{center}
\centerline{\includegraphics[width=0.478\columnwidth]{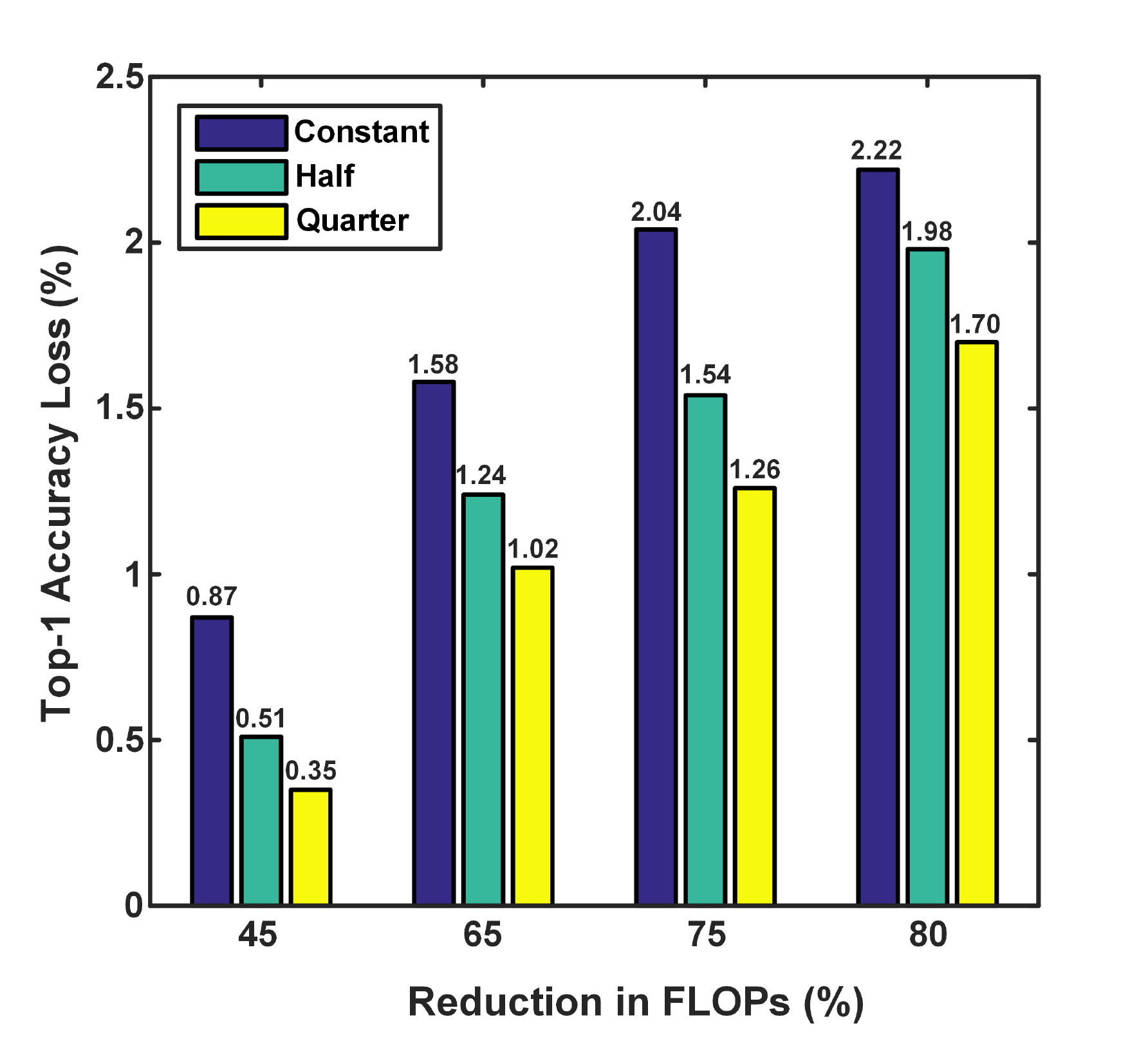}}
\caption{Comparison results of three compression degrees with network depth for compression of ResNet34 on ImageNet}
\label{figure5}
\end{center}
\end{figure}

\begin{figure} [t]
\centering
\subfigure[Conv3\_3]{
\begin{minipage}{5.6cm}
\centering
\includegraphics[scale=0.32]{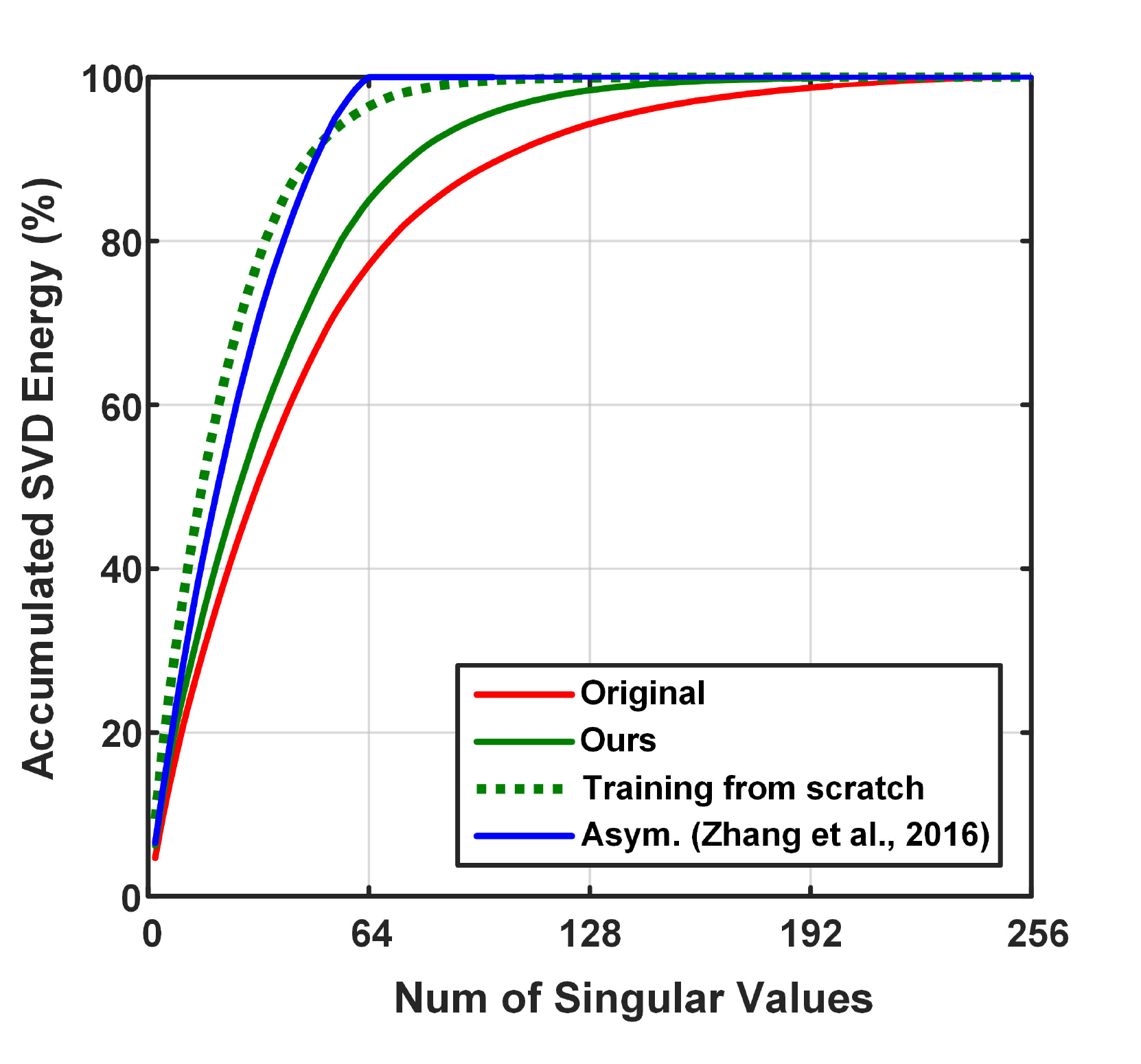}
\end{minipage}
}
\subfigure[Conv5\_3]{                    % 第二张子图
\begin{minipage}{5.6cm}
\centering                                                          %子图居中
\includegraphics[scale=0.32]{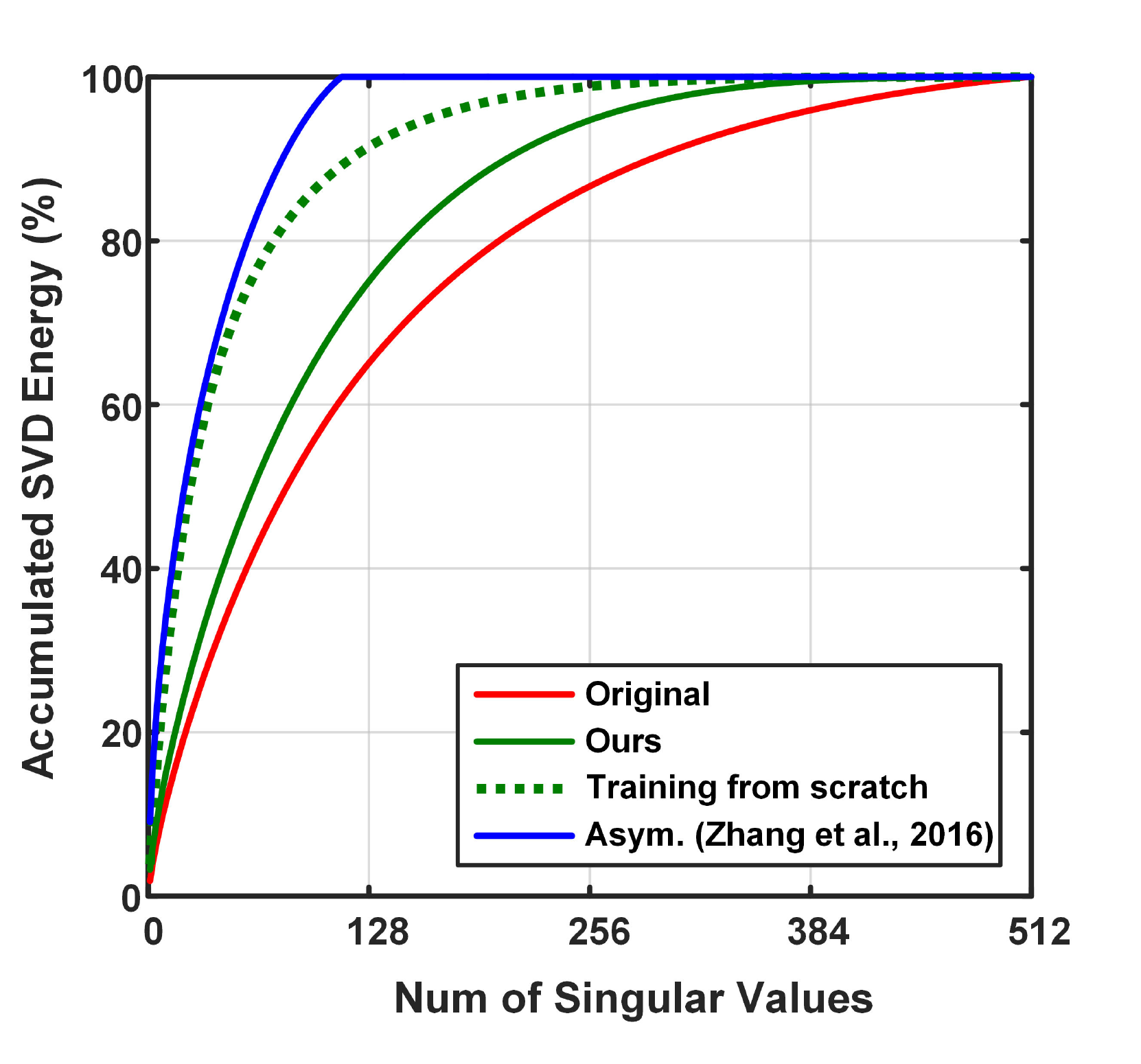}                %以pic.jpg的0.5倍大小输出
\end{minipage}
}
\caption{SVD accumulative energy curves of Conv3\_3 and Conv5\_3 in VGG16 and its compressed networks. The red solid line denotes the original network; the green solid line denotes the compressed given by our method; the green dot line denotes training from scratch with the same architecture as our compressed model} %                         % 大图名称
\label{figure6}                                                        %图片引用标记
\end{figure}

\subsection{Compression Degree with Network Depth}
As mentioned in Section~\ref{degree}, the ratio of $n$ between adjacent stages is $1 : 4$ (hereinafter called \emph{`Quarter'}) in our experiments. In this part, we consider another two degrees of compression.
The first one is \emph{`Constant'}, which means the \emph{n} for all stages maintain the same. The second one is \emph{`Half'}, which means the ratio of \emph{n} between adjacent stages is $1:2$.
We evaluate these three kinds of strategies on compression of ResNet34.
Fig.~\ref{figure5} illustrates the results. Compressed networks with \emph{`Quarter'} degree give the best performance, which is consistent with \cite{Ioannou2016Deep}. These results also indicate that there should be less compression with increasing depth.

%The compression ratio of our method is controlled by $n$. In each compressed model, we set a same $n$ for all layer in above experiments. This is unthoughtful since different layers are not equally redundant. \cite{Ioannou2016Deep} discussed the impact of the group numbers in group convolutional layers with network depth. A \emph{root} architecture, whose group numbers are decreased with depth (\emph{e.g.} 8-4-2-1), was proposed with better performance than \emph{column} (constant group numbers with depth, \emph{e.g.} 4-4-4-4) and \emph{tree} (increased group numbers with depth, \emph{e.g.} 1-2-4-8) architectures, which indicated less redundancy in later layers. Taking the point \cite{Ioannou2016Deep} mentioned into consideration, we further decompose the original network into the form of the \emph{root} architecture. Fig.~\ref{figure5} illustrates the results on ResNet34. Compressed models with \emph{root} architectures give better performances than those with \emph{tree} architectures. Under $4\times$ compression ratio (75.2\% reduction of FLOPs), the \emph{root} architecture increases the error rate by 1.26\%, and under $5\times$ (80.4\% reduction of FLOPs) the accuracy loss of \emph{root} architecture is 1.70\%.

\begin{figure}[t]
\begin{center}
\centerline{\includegraphics[width=0.5\columnwidth]{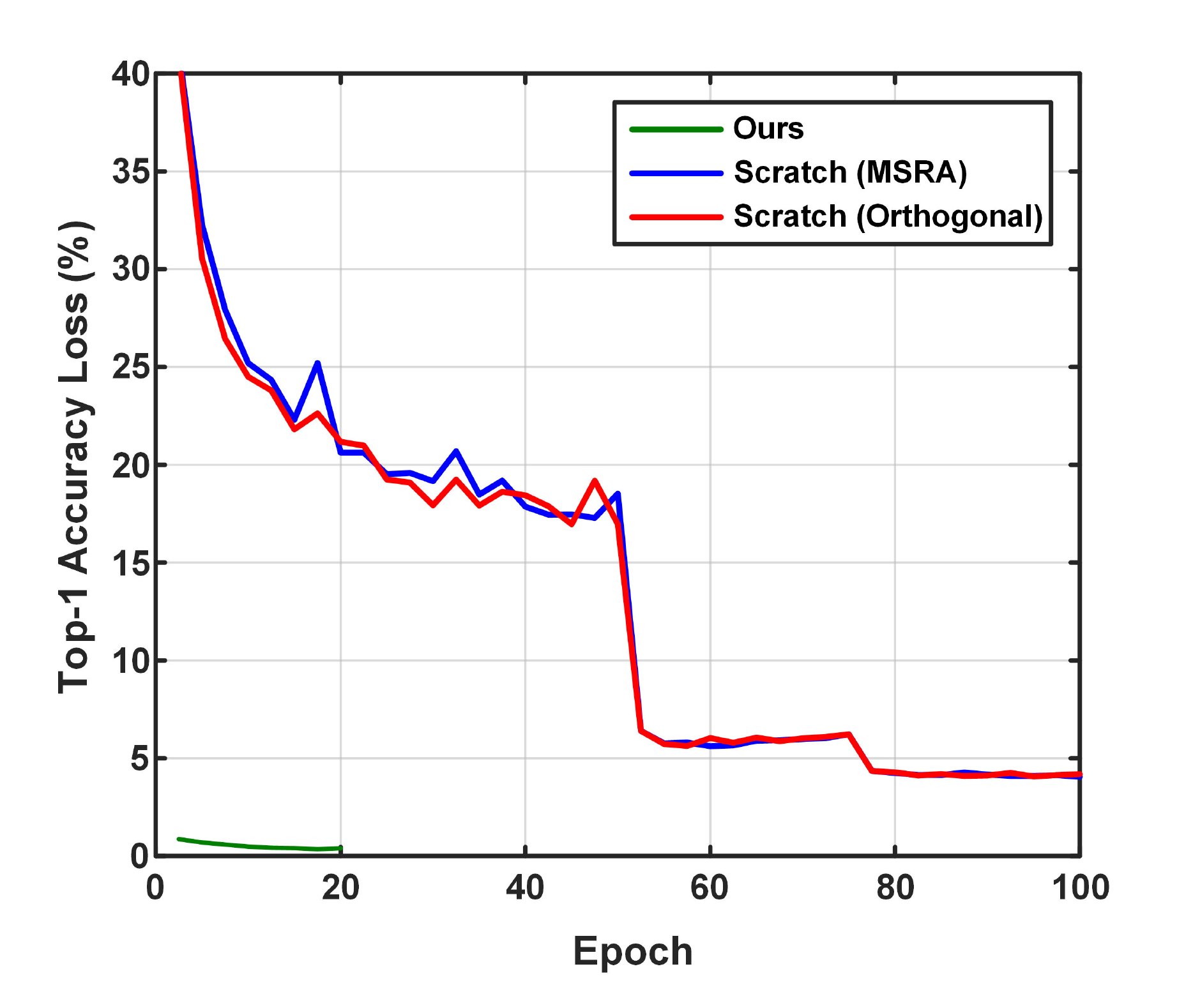}}
\caption{Learning curves of compressed networks with initializations of ours, `MSRA' and `Orthogonal'}
\label{figure7}
\end{center}
\end{figure}

%\subsection{Discussion and Analysis}
%Next, we will further provide discussions on our method in two aspects.

\subsection{Degeneracy of the Extremely Compressed Network}
Fine-tuning is a necessary process to recover the performance of a compressed network. However fine-tuning after low-rank decomposition may suffer from the instability problem \cite{Lebedev2015Speeding}. In our experiments for VGG16 on ImageNet, the training of compressed models from \emph{Asym.} shows gradient explosion problems, which was also mentioned in \cite{Zhang2016Accelerating}. Besides, the performance degraded quickly when the compression ratio was getting larger.
This phenomenon is caused by the degeneracy of the compressed network \cite{Orhan2017Skip,Saxe2013Exact,Xie2017All}.

As we analyzed in Section~\ref{degeneracy}, we calculate the singular values of Jacobian to analysis the degeneracy problem. Fig.~\ref{figure6} illustrates SVD accumulative curves of two layers from the original VGG16 and compressed networks ($5\times$ speed-up) from \emph{Asym.}, our method and training from scratch with the same architecture as our compressed model. The curves of the original network (denotes as red solid lines) are the most flat, indicate the least degenerate. Ours (green solid) are the second, training from scratch (green dotted) and \emph{Asym.} (blue solid) are worse than ours. The rest layers hold similar phenomena. Thus it can be concluded that, our proposed method can alleviate the degeneracy problem efficiently, while \emph{Asym.} is much affected due to more elimination of singular value. The result also proves that training from scratch can not provide enough dynamic to conquer the problem of degeneracy~\cite{Xie2017All}.

The problems of training instability are lightened in ResNet due to the existence of Batch Normalization~\cite{Ioffe2015Batch} and short-cut connection \cite{Orhan2017Skip}. However the performance of \emph{Asym.} on ResNet still degraded quickly as the compression ratio increased. The degeneracy of network is still an open problem. We believe that it should be taken into account when studying network compression. %\cite{Xie2017All} explored a better solution for degeneracy with orthonormality and modulation, which inspires us a lot, and we will focus on this direction in our future work.

\subsection{A Better Initialization of the Network}
The comparison between training from scratch and our approximation method indicates that our compressed models provided a better initialization. For a further verification, we also evaluate the orthogonal initialization which could alleviate degeneracy in linear neural networks \cite{Saxe2013Exact}. We compare models trained with the architecture of \emph{`Ours-Res34/A'}. Fig.~\ref{figure7} illustrates the learning curves. Our compressed network achieves much lower accuracy loss.

\begin{figure}[t]
\centering
\subfigure[Ours]{
\begin{minipage}{3.45cm}
\centering
\includegraphics[scale=0.36]{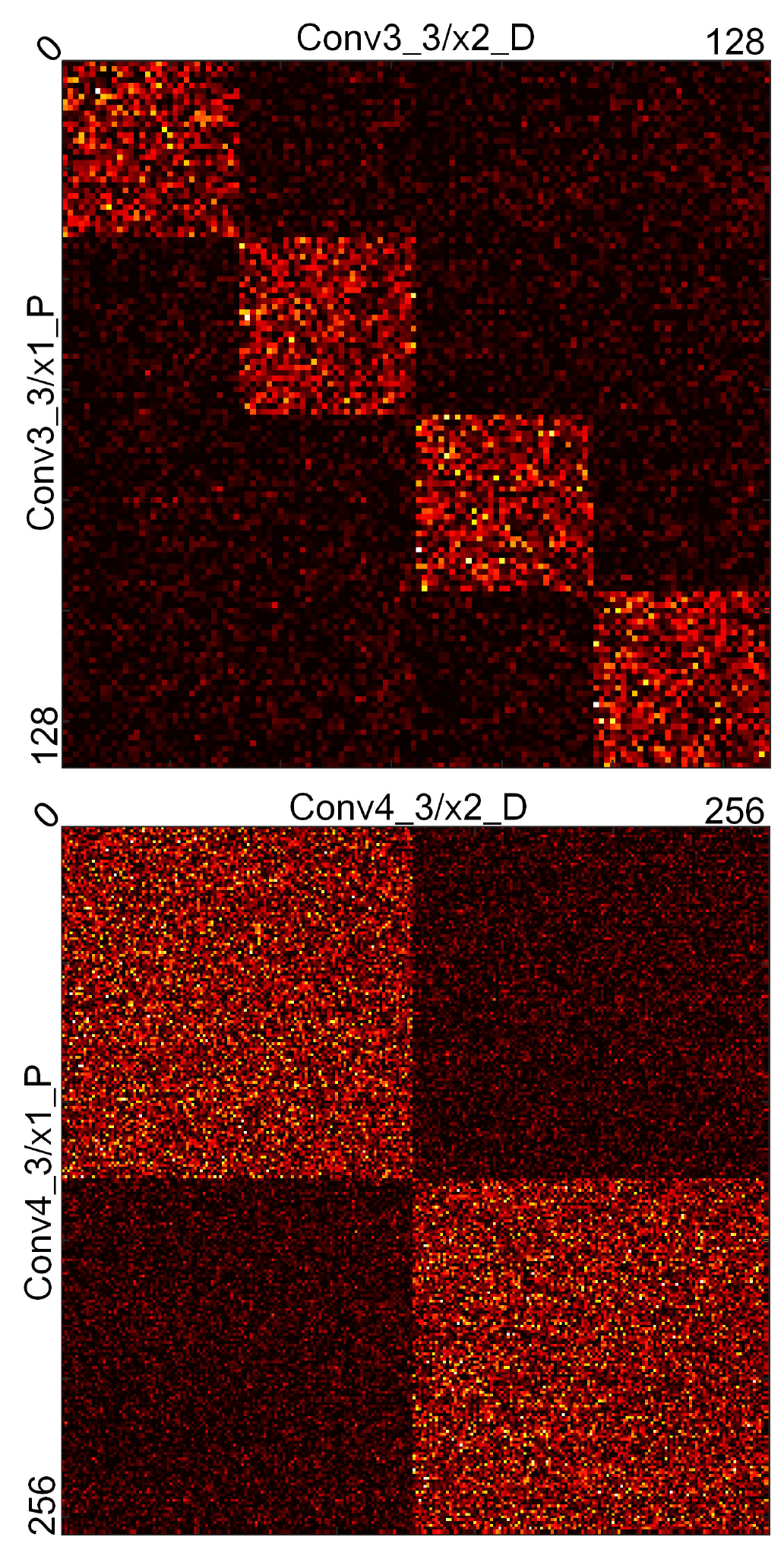}
\label{figure8:a}
\end{minipage}
}
\subfigure[Scratch (MSRA)]{
\begin{minipage}{3.45cm}
\centering
\includegraphics[scale=0.36]{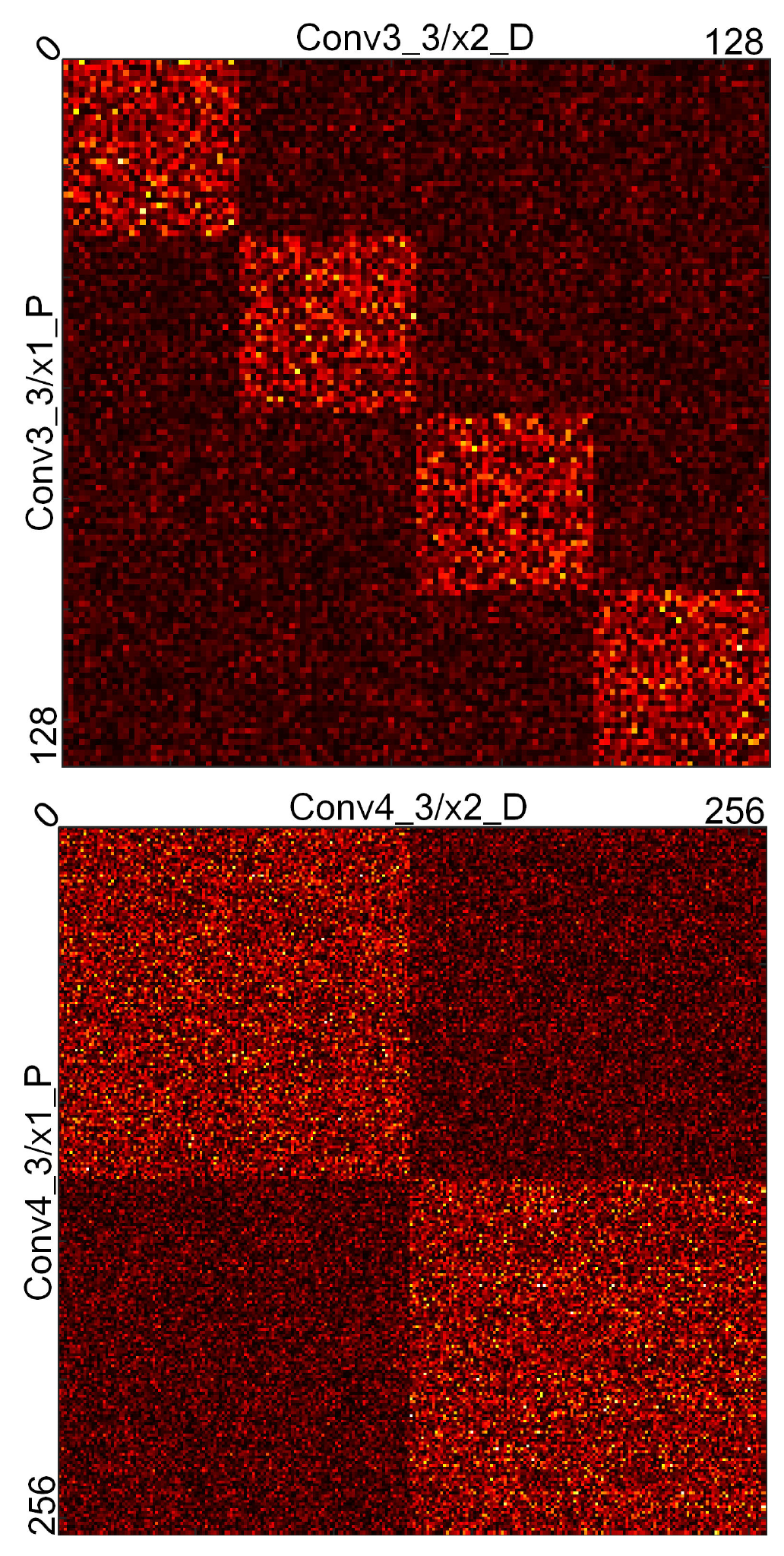}
\label{figure8:b}
\end{minipage}
}
\subfigure[Scratch (Orthogonal)]{                    % 第二张子图
\begin{minipage}{4.45cm}
\centering                                                          %子图居中
\includegraphics[scale=0.36]{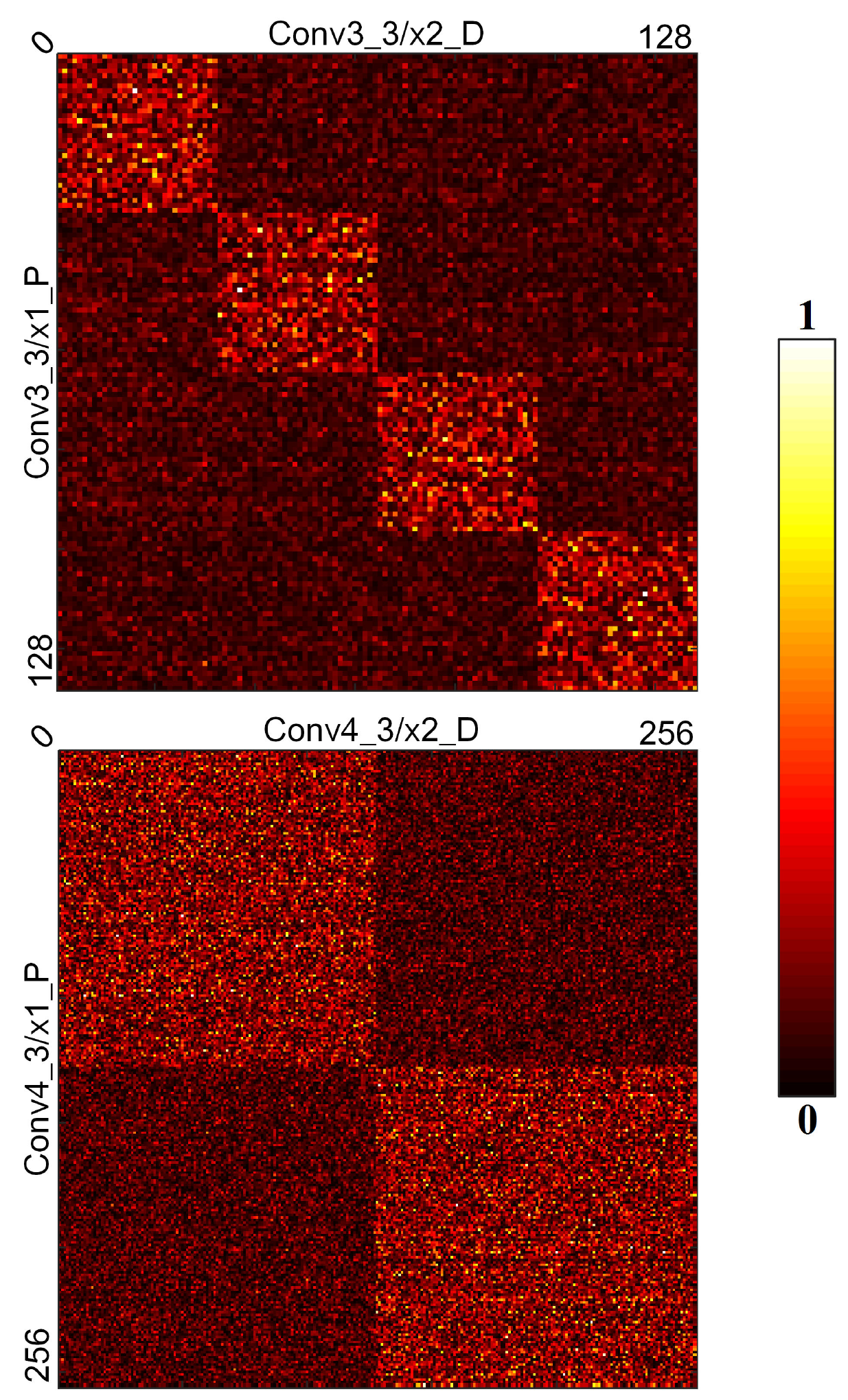}                %以pic.jpg的0.5倍大小输出
\label{figure8:c}
\end{minipage}
}
\caption{Inter-layer filter correlation using initializations of ours, `MSRA' and `Orthogonal'. Brighter pixel indicates higher correlation} %                         % 大图名称
\label{figure8}                                                        %图片引用标记
\end{figure}

To give a better understanding, we conduct the inter-layer filter correlation analyses \cite{Ioannou2016Deep}. Fig.~\ref{figure8} shows the correlation between the output feature maps of filter group convolution and its previous point-wise convolution. Enforced by the filter group structure, the correlation shows a block-diagonal matrix. Pixels in the block-diagonal part of ours (Fig.~\ref{figure8:a}) are brighter than those of training from scratch (Fig.~\ref{figure8:b} and~\ref{figure8:c}), and reversed in the background. It means features between groups are more independent in our method. Similar phenomena are observed in other layers, which confirm that our proposed method gives a better initialization of the network.

\section{Conclusion}
In this paper, we proposed a filter group approximation method to compress networks efficiently. Instead of compressing the spatial size of filters or the number of output channels in each layer, our method is aimed at exploiting a filter group structure of each layer. The experimental results demonstrated that our proposed method can achieve extreme compression ratio with tiny loss in accuracy. More importantly, our method can efficiently alleviate the degeneracy of the compressed networks. In the future, we will focus on solving the degeneracy problem on network compression.

\bibliographystyle{splncs04}
\bibliography{2102}
\end{document}